# An Empirical Evaluation of Using ChatGPT to Summarize Disputes for Recommending Similar Labor and Employment Cases in Chinese

Po-Hsien Wu[‡]     Chao-Lin Liu[†]     Wei-Jie Li[ǁ]
National Chengchi University, Taiwan
{chaolin[†],111753120[‡],110753128[ǁ]}@nccu.edu.tw

**Abstract.** We present a hybrid mechanism for recommending similar cases of labor and employment litigations. The classifier determines the similarity based on the itemized disputes of the two cases, that the courts prepared. We cluster the disputes, compute the cosine similarity between the disputes, and use the results as the features for the classification tasks. Experimental results indicate that this hybrid approach outperformed our previous system, which considered only the information about the clusters of the disputes. We replaced the disputes that were prepared by the courts with the itemized disputes that were generated by GPT-3.5 and GPT-4, and repeated the same experiments. Using the disputes generated by GPT-4 led to better results. Although our classifier did not perform as well when using the disputes that the ChatGPT generated, the results were satisfactory. Hence, we hope that the future large-language models will become practically useful.

## 1    Introduction

Searching and finding similar previous cases are the basis for various possible applications in the legal domain. Similar previous cases may provide clues about how judges may judge a new case and what lawyers can do in oral arguments and cross examinations [4][7]. The research for searching similar previous cases started many years ago, and has attracted the attention of many researchers in recent years.

We will mention some closely related issues now, but we cannot offer a comprehensive survey for the research of recommending similar previous cases. Researchers have explored two possible research directions. Some suggested that the citation-network-based method may be less applicable to the legal domain [5], and we and many others took the text-based approach [11]. In the text-based camp, researchers must determine how they define the similarity between two cases. This includes at least two issues. First, how do we define similarity between two cases numerically? Second, do we compare every statement in the two texts as a whole, or do we compare selected parts partially [11]? With the support from advanced artificial intelligence techniques, it is possible to consider more specific background information about the lawsuits in comparing the cases [2]. Low-resource remains a challenge for similar case recommendation research. CAIL-2019 [18], which represents the national corpus of China, contains slightly more than 8000 labeled pairs of cases [2][3].

In previous work, we explored using a clustering-based method for recommending similar cases [10]. Two cases were judged to be similar if they had similar distributions in the clusters of disputes. Hence, we have a narrow but practical standard of similarity. The disputes between the employees and the employers are the main source of the litigations. The scale of our experiments was thus limited by the number of cases that explicitly listed disputes between the litigants. In this paper, we report the results of our classification of whether or not two cases are similar. We built the classifiers with deep-learning approaches, and judged whether the two given cases were similar based on the disputes between the litigants of civil cases, i.e., the labor and employment cases. This can be achieved with our previous dataset.

With the availability of ChatGPT, we evaluated the disputes that we asked the ChatGPT to generate from the litigants' claims. If ChatGPT would serve as a reliable source of itemized disputes, we could expand the scale of our experiments [19]. We replaced the GPT-generated disputes with the original disputes in our classification experiments. If the new outcomes are satisfactory, we may consider using ChatGPT to summarize the disputes of the litigants for cases that do not have itemized disputes, thus alleviating the problems of low resources for our approach.

We observed encouraging results when using GPT-3.5 and GPT-4. In the remainder of this paper, we offer a formal definition of our research problem in Section 2. We provide more background information about our data in Section 3, and explain how we used ChatGPT to summarize the litigants' claims for us in Section 4. We elaborate on the design of the experiments in Section 5, and report the experimental results in Section 6. We then wrap up with some discussions in Section 7.

## 2    Problem Definition

We evaluate the potential contribution of ChatGPT in comparing the similarity of two labor and employment cases in Chinese.

In a previous work [10], we evaluated clustering-based methods for recommending similar labor and employment cases. Since we have labeled some case pairs by their similarity, we could train some classifiers with these annotated cases, and use the trained classifiers to guess whether two future cases are similar.

The real challenge is how we and the annotators could have determined whether the two cases were similar. It is not easy to define "similar" even in everyday life, let alone in judicial cases. In our work, we focused on whether the disputes between the employees and employers in the two cases are similar. This is certainly not the only way to define "similarity" between two cases. For instance, one might be more interested in which party won the cases or in the industry and years of the cases.

Based on the narrowed perspective of similarity, we can define our work in the following way. Assume that we have collected the judgment documents of $m$ previous cases that explicitly included the disputes between the employees and the employers. We denote this collection as $\boldsymbol{C} = \{C_1, C_2, \cdots, C_m\}$. Let $\boldsymbol{K} = \{K_1, K_2, \cdots, K_m\}$, where $K_i$ denote the descriptions of disputes listed in the judgment documents for case $C_i$. We annotated a collection of case pairs, $C_i$ and $C_j$, $i \neq j$, with their degrees of similarity $S_{i,j}$, where $S_{i,j}$ can a label in $S$ = {"similar", "barely similar", "not similar"}. We denote the collection of annotated case pairs and their similarity as $D = \{(C_i, C_j, S_{i,j}) | C_i \in C, C_j \in C, S_{i,j} \in S\}$. Assume that we have $n$ pairs of labeled cases. We may denote the annotated collection as $D = \{D_1, D_2, \cdots, D_n\}$. Namely, a $D_k$ represent a $(C_i, C_j, S_{i,j})$ in $D$ for certain $i$ and $j$.

We split $D$ into two mutually exclusive subsets, one for training and one for testing. We denote the training and testing subsets as $R$ and $E$, respectively, and $D = R \cup E$. We trained classifiers with machine learning methods and with instances in $R$, and evaluated the quality of the trained classifiers with the instances in $E$. Let $f$ denote the classification function of a trained classifier. When given a case pair, $f$ will return the similarity of the case pair. More specifically, if $(C_i, C_j, S_{i,j}) \in E$, we hope that $f(C_i, C_j) \equiv f(K_i, K_j) = S_{i,j}$.

## 3  Data Source, Selection and Preprocessing

We provide information about the data sources, discuss some basic statistics of our data, and introduce an important step in the preprocessing procedure in this section. We have described the major steps for data preparation in previous papers in JURIX [8] and JURISIN [9] and the legalAIIA workshop in ICAIL [10]. We will not repeat all of the details to avoid the concerns of self-plagiarism.

### 3.1  Data Source: Taiwan Judicial Yuan

We obtained the judicial documents from an open repository that is maintained by the Judicial Yuan. The Judicial Yuan, the highest governing body for Taiwan's judicial system, oversees the publication of judgment documents from various courts, including local, high, supreme, and special courts. These documents are typically released on the TWJY website three months after the judgment date, with February judgments, for example, becoming available in May. Some documents may not be published due to legal reasons, such as protecting minors or litigants, and their contents are anonymized for privacy.

As of July 2023, the TWJY website hosts approximately 18.7 million documents, dating back to January 1996. Initially, only documents from a limited number of special courts were available in the first few years, with broader coverage starting from 2000 onward. The Judicial Yuan updates the website monthly, with a three-month lag, ensuring that users can access and download new judgment documents from various courts in a compressed file. However, the number of available documents may fluctuate due to retractions based on legal reasons. Anonymization of published documents is mandated by law, and the government takes responsibility for safeguarding the privacy of individuals involved in lawsuits.

## 3.2 Data Selection: Cases with Listed Disputes

Each document in the TWJY is a JSON file and adopts a common top-level structure. The structure consists of seven fields: JID is the long identification number; JYEAR is the year when the case occurred in terms of Taiwan calendar; JCASE is the abbreviated code for the type of the lawsuit; JNO is the short identification number for the lawsuit; JDATE is the date for the current judgment in terms of the Western calendar; JTITLE is the category of the lawsuit, and JFULL is the full text for the judgment document.

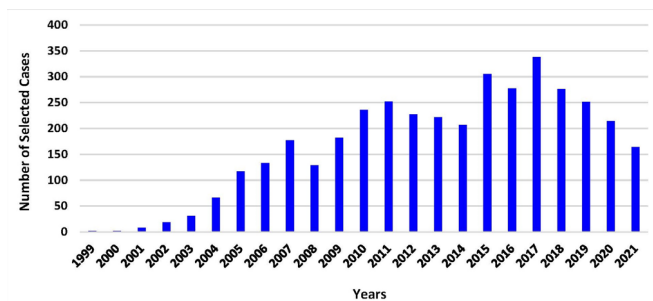

**Figure 1**: The temporal distribution of the selected cases [10]

Therefore, using the contents of JCASE and JTITLE fields to find judgment documents of labor and employment litigations, among a myriad of case categories, is a basic step. We focus on the judgments of the local courts, where the judges would consider the factual parts of the lawsuits, and using the codes in the JCASE and the JTITLE fields could help us exclude appeal cases.

The filtering of relevant and usable documents needs more steps. Sometimes, even when the JCASE and JTITLE fields seem to qualify a document, we may find clues in the JFULL field that indicate the case does not meet our needs.

Most importantly, in the current study, we look for cases in which the courts explicitly recorded the disputes between the plaintiffs and the defendants in the judgment document. We look into the JFULL field to ensure the documents meet this requirement. The listed disputes provide important information about the lawsuits, and help the lawsuits to proceed more effectively. Nevertheless, not all of the judgment documents would record the disputes.

At this moment, we found 3835 cases from 21 local courts in Taiwan. Figure 1 shows the distribution over the years when the cases took place. The horizontal axis shows the years, and the vertical axis shows the number of cases. The long-term trend is that the number of cases has increased. In addition, most of the selected cases came from the top five courts, which happened to be located in metropolitan or industrialized areas [10].

## 3.3 Disputes in Labor and Employment Cases

For each of the judgment documents we selected in Section 3.2, there is a section that itemizes the disputes between the plaintiffs and the defendants. The format looks like the following, although the exact formats may vary, and they are actually in Chinese. (See Appendix A for real examples.)

> **The disputes are as follows.**
>  **dispute-1 statement**
>  **dispute-2 statement**
>  **…**

The challenge is that the statements contain specific information about their belonging cases, e.g., person names, place names, and time expressions. Comparing these named entities between two litigations may not make very much sense.

For this reason, our programs would recognize using the NER techniques (named entity recognition) to identify the named entities. They would replace those specific nouns (or noun phrases) with more general terms, e.g., someone, somewhere, and some time. This would make the dispute statements more comparable. We refer to this as a "blurring" step. (See Appendix A for real examples.)

## 4 Summarizing Disputes with ChatGPT

Large language models and generative AI have shown their impressive applicability in a wide range of tasks in the past year, including tasks that are in the legal domain [16]. ChatGPT of OpenAI is perhaps the most famous leader for this ground-breaking development. One main concern of applying the generative technology was the hallucination problem. There is a case in which a legal practitioner used the ChatGPT to generate a legal document that cited a non-existing document [15]. The research community has been very concerned about this annoying problem very intensively.

As users, we can avoid the hallucination problem using prompts with specific constraints. In a typical judgment document, we can find statements of the plaintiffs and the statements of the defendants. A list of disputes between the plaintiffs and the defendants will follow, if the document has the list. We extracted and used the statements of the plaintiffs and the statements of the defendants from the judgment documents in our prompts.

Since our goal was to ask the ChatGPT to summarize the disputes between the plaintiffs and the defendants, we must try to confine the ChatGPT to finding the disputes only from the statements of the plaintiffs and the defendants. To this end, our prompts consist of three steps. First, we ask ChatGPT to summarize the plaintiffs' statements with a list of items. Second, we ask the ChatGPT to summarize the defendants' statements with a list of the items. Then, we asked ChatGPT to itemize the disputes between the plaintiffs and the defendants based on the two lists it returned in the previous two steps. Upon the request of the reviewers, we provide the segment of our Python code for this three-step operation in Appendix B.

The disputes that we received would look like the examples that we explained in Section 3.3, and we could use them in our experiments as if they were the disputes that were listed in the original judgment documents.

We have implemented the ideas with GPT-3.5 and GPT-4 (0613). We set the parameters temperature to 0.7 and 0.3 for GPT-3.5 and GPT-4, respectively, to avoid overgeneration. Currently, both ChatGPT versions limit the number of tokens for the conversations, so we confine the number of tokens of our prompts to 11500 and 6000 for

GPT-3.5 and GPT-4, respectively. We would drop the cases if the lengths

| Source | Cases | Sent. | Labeled Pairs | Labeled Cases |
|---|---|---|---|---|
| TWJY | 3835 | 12385 | 2288 (1360, 928) | 1543 |
| GPT-3.5 | 3771 | 14278 | 2277 (1355, 922) | 1541 |
| GPT-4.0 | 3826 | 16878 | 2288 (1360, 928) | 1543 |

**Table 1.** Basic statistics of the source corpora.

of the statements of the plaintiffs or the defendants exceeded these limits.

In addition, we might not receive the summaries or the disputes from the ChatGPT sometimes. When this unpredictable problem occurred, we resubmitted the prompts at most three times. If the problem persisted, we would drop the case as well. As a result, we started with 3835 judgment documents (cf. Section 3.2), and obtained disputes of 3771 and 3826 documents from GPT-3.5 and GPT-4, respectively.

## 5   Classification for Similar Case Recommendations

We reported some results of applying clustering-based methods to recommend similar cases in [10]. In this presentation, we reported the results of using classification-based methods for the recommendation task. We will report the resulting differences in the quality of the recommendations in Section 6.

### 5.1   Data for Training, Validation, and Test

Table 1 provides more statistics about the judgment documents used in the current work. We have 12,385 dispute statements in the selected cases from the TWJY. We have annotated 2288 pairs of similar cases: 1360 not-similar pairs and 928 similar pairs [10]. Only 1543 of the 3835 cases were part of the labeled pairs of cases. We did not have 2×3835=7670 labeled cases because some labeled pairs shared a common case.

Since all of these 1543 cases were not used in finetuning the BERT models (cf. Section 5.2), there would not be data leakage problems for embedding. In addition, we used cosine similarity in the classifiers, so using the same case in different pairs in the training and test would not cause the problem of data leakage.

Due to the limitations of GPT-3.5, we did not obtain the dispute statements for cases when we used GPT-3.5 (cf. Section 4). Hence, we had fewer labeled pairs and cases when we used GPT-3.5 in our experiments.

### 5.2   Text Embedding and Fine-Tuning the Pretrained BERT models

We vectorized the statements of disputes only with the TFIDFVectorizer of scikit-learn [10]. Now, we vectorized the statements with the Sentence-BERT [13] with either the Lawformer [17] or with the Chinese RoBERTa [1].

The BERT models are both for Chinese, only that the Lawformer is specifically pretrained with legal documents of China that were recorded with simplified Chinese. Hence, in our experiments, we may choose to fine-tune the BERT models with the legal documents of Taiwan, which were recorded in traditional Chinese. There are some variations between the simplified and traditional Chinese and intricate differences between the legal terms used in China and Taiwan.

We fine-tuned the Lawformer and RoBERTa along with the Sentence-BERT. We first put aside the sentences in the 1543 labeled cases from the sentences in the 3835 TWJY cases for training and testing our classifiers. We then clustered the remaining 5321 sentences with a typical density-based clustering method, and we disregarded all the sentences in small clusters that contained less than ten sentences. The result is 3031 sentences.

We created pairs of sentences for fine-tuning the Sentence-BERT as follows. First, we created pairs of sentences that belonged to the same clusters, and assigned this type of pairs to the category "same". Then, we created pairs of sentences that belonged to different clusters, and assigned this type of pairs to the category "diff". Ultimately, we created a database of 3875115 "diff" pairs and 716850 "same" pairs. We randomly sample only 50000 from each category to fine-tune the Sentence-BERT for efficiency and effectiveness.

### 5.3 Transforming the Text for Convolutional Neural Networks

Without loss of generality, we will consider the task to determine whether two cases $C_i$ and $C_j$ are similar in the following deliberation. Following the notation that we introduced in Section 2, we compute the cosine similarity between the disputes in $K_i$ and $K_j$. Let $K_i = \{k_{i,1}, k_{i,2}, \cdots, k_{i,\alpha}, \cdots, k_{i,u}\}$ and $K_j = \{k_{j,1}, k_{j,2}, \cdots, k_{j,\beta}, \cdots, k_{j,v}\}$. The cosine similarity $s_{\alpha,\beta}^{i,j}$ between a given pair of disputes, $k_{i,\alpha}$ from $K_i$ and $k_{j,\beta}$ from $K_j$, is defined in (1), where $v_{i,\alpha}$ and $v_{j,\beta}$ denote the Sentence-BERT vectors of $k_{i,\alpha}$ and $k_{j,\beta}$, respectively. Theoretically, the range of $s_{\alpha,\beta}^{i,j}$ is $[-1,1]$, but the majority were in the range of $[0,1]$, and $s_{\alpha,\beta}^{i,j} = 1$ only when $k_{i,\alpha} = k_{j,\beta}$.

$$s_{\alpha,\beta}^{i,j} = cosine\_similarity(v_{i,\alpha}, v_{j,\beta}) \qquad (1)$$

We construct a matrix $M_{i,j}$ for $C_i$ and $C_j$, and an element $m_{\alpha,\beta}^{i,j}$ in $M_{i,j}$ is set to $1 - s_{\alpha,\beta}^{i,j}$. Namely, the element $m_{\alpha,\beta}^{i,j}$ in $M_{i,j}$ at the position $(\alpha, \beta)$ is the similarity between the $\alpha$th and $\beta$th disputes of $C_i$ and $C_j$, respectively.

We still clustered all of the disputes of the cases in $C$, as we did and explained in [10], except that we switched to using the HDBSCAN of scikit-learn for clustering.[1] Let $m_{max}^{i,j}$ and $m_{min}^{i,j}$ denote the largest and the smallest values in $M_{i,j}$, respectively. We set the parameter cluster_select_epison, $\epsilon$, for HDBSCAN to $m_{min}^{i,j} + 0.8 \times (m_{max}^{i,j} - m_{min}^{i,j})$ to control the number of clusters that HDBSCAN may produce. Let $\mathbb{C} = \{c_1, c_2, \cdots, c_\gamma\}$ denote the resulting clusters that contained all of the disputes in $K$. The actual values of $\gamma$ varied in different experiments and were determined automatically by HDBSCAN. For convenience of communication, we set the elements in $\mathbb{C}$ to integers. Namely, $c_1 = 1$, $c_2 = 2$, …, and $c_\gamma = \gamma$. After the clustering step, each dispute $k_{i,\alpha}$ in a $K_i \in \mathbf{K}$ will be assigned to a certain cluster code $c_{i,\alpha} = c \in \mathbb{C}$. Let $\aleph = \{\aleph_1, \aleph_2, \cdots, \aleph_m\}$, where $\aleph_i = \{c_{i,1}, c_{i,2}, \cdots, c_{i,|K_i|}\}$ and $|K_i|$ is the number of disputes in $K_i$. We evaluated the idea that whether we may judge two cases $C_i$ and $C_j$ were similar given the size of $\aleph_i \cap \aleph_j$ in [10].

---

[1] https://scikit-learn.org/stable/modules/generated/sklearn.cluster.HDBSCAN.html

We then created a square matrix $Z^{i,j}$ for $C_i$ and $C_j$ in the following way. Notice that we let $K_i = \{k_{i,1}, k_{i,2}, \cdots, k_{i,\alpha}, \cdots, k_{i,u}\}$ and that each $k_{i,\alpha} \in K_i$ has been assigned a cluster code $c_{i,\alpha}$. Hence, we could reorder the disputes in $K_i$ by their cluster codes. Let $K_i'$ in (2) denote such a reordered version of $K_i$. We reordered the disputes in $K_j = \{k_{j,1}, k_{j,2}, \cdots, k_{j,\beta}, \cdots, k_{j,v}\}$ as well, so we have the reordered $K_j'$ in (3).

$$K_i' = \{k_{i,1}', k_{i,2}', \cdots, k_{i,\alpha}', \cdots, k_{i,u}'\} \tag{2}$$
$$K_j' = \{k_{j,1}', k_{j,2}', \cdots, k_{j,\beta}', \cdots, k_{j,v}'\} \tag{3}$$

We concatenate $K_i'$ and $K_j'$ to form $K^{i,j}$ as shown in (4). By this concatenation step, we could create a squared matrix that would facilitate our operations with the convolutional neural networks. The value of the element $z_{x,y}^{i,j}$ in $Z^{i,j}$ at position $(x, y)$ is the cosine similarity between the $x$th and $y$th elements in $K^{i,j}$, counting from the left to the right. Therefore, the size of $Z^{i,j}$ is $(|K_i| + |K_j|) \times (|K_i| + |K_j|)$, in general.

$$K^{i,j} = \{k_{i,1}', k_{i,2}', \cdots, k_{i,u}', k_{j,1}', k_{j,2}', \cdots, k_{j,v}'\} \tag{4}$$

Finally, let $z_{max}^{i,j}$ and $z_{min}^{i,j}$ denote the largest and the smallest values in $Z^{i,j}$, respectively. We projected a value of $z_{x,y}^{i,j}$ in $Z^{i,j}$ to the range $[0, 255]$ with formula (5). This step practically created a squared grey-level image. Since the selected $\epsilon$ is closer to $z_{maz}^{i,j}$ than to the average of $z_{min}^{i,j}$ and $z_{maz}^{i,j}$, giving a half of $[0, 255]$, i.e., $[128, 255]$, to the range of $[\epsilon, z_{maz}^{i,j}]$ is to monitor the range $[\epsilon, z_{maz}^{i,j}]$ with a higher resolution.

$$\begin{aligned} z_{x,y}^{i,j} &\leftarrow \frac{z_{x,y}^{i,j} - z_{min}^{i,j}}{\epsilon - z_{min}^{i,j}} \times 127, if\ z_{x,y}^{i,j} \leq \epsilon \\ z_{x,y}^{i,j} &\leftarrow 127 + \frac{z_{x,y}^{i,j} - \epsilon}{z_{maz}^{i,j} - \epsilon} \times 128, if\ z_{x,y}^{i,j} > \epsilon \end{aligned} \tag{5}$$

We would evaluate whether the grey-level images would be useful for predicting whether or not cases $C_i$ and $C_j$ are similar. The reordering operations, i.e., the steps to create (2) and (3), allow us to figure out the clusters that included more dispute statements, thereby creating contextual information. Figure 2 shows a sample image for similar and a sample image for dissimilar case pairs. (The bright diagonal from the upper left to the lower right corners was a natural result of disputes being compared with themselves.)

### 5.4 Training the Classifier

Figure 3 shows the flow of using a simple procedure with the convolutional neural network (CNN) layers and the max-pooling layers for the classification task. We scaled a given $Z^{i,j}$ to the size of $32 \times 32$. The selection of 32 was based on the number of disputes listed in the judgment documents. This image would go through two iterations of CNN and max-pooling. In each iteration, we used 32 CNN units, and the convolution mask was $3 \times 3$. We used a drop-out layer of 0.2 in between the CNN and the max-pooling layers. Then, the image would be shrunk by the $2 \times 2$ max-pooling step. Then, after the two iterations, an $8 \times 8$ matrix would be flattened and used as the input to a fully connected layer to produce the classification results.

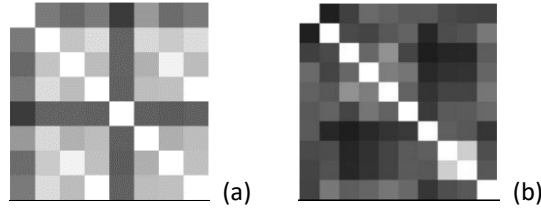

**Figure 2**: The grey-level images offer a special perspective for the similarity between the cases: (a) a similar pair (b) a dissimilar pair

We trained the classifier with the data that we listed in Table 1. We used 64% of the labeled pairs for training and 16% for validating the classification models. We reserved 20% for the final test. We split the data with a stratification process.

## 6 Empirical Evaluations

We conducted experiments to answer two questions. The first is whether we can use ChatGPT to summarize and itemize the disputes for the labor and employment litigations. The second is to examine whether or not and how to use the Lawformer in our studies. More specifically, should we use an ordinary Chinese RoBERTa or Lawformer for embedding? What if we fine-tune both BERT models?

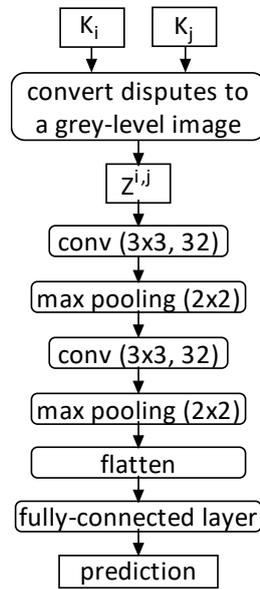

**Figure 3**: The CNN classifier

Since random numbers play important roles in the operations of artificial neural networks (deep learning), we repeated each of our experiments 30 times to draw a boxplot for the experiment. The training, validation, and test data were resampled every time. Since the BERT models were fine-tuned with data that were completely different from the data that we listed in Table 1, we did not repeat the finetuning step.

After completing the training process, we asked the classifiers to predict whether or not the pairs of cases in the reserved 20% of data were similar. We only fine-tuned our BERT model once, but re-split the labeled pairs of cases 30 times so that we could make boxplots for the results. We can create boxplots for both the F1 measure and the traditional accuracy.

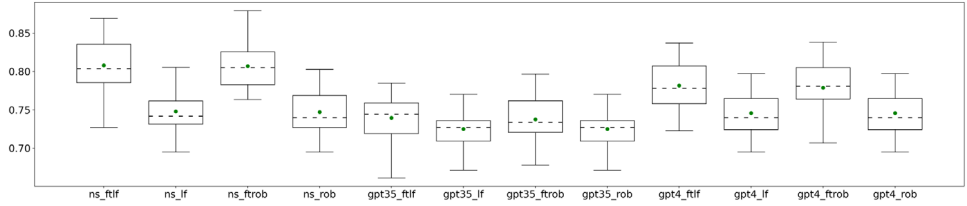
**Figure 4**: F$_1$ measures of the classification with the CNN classifier

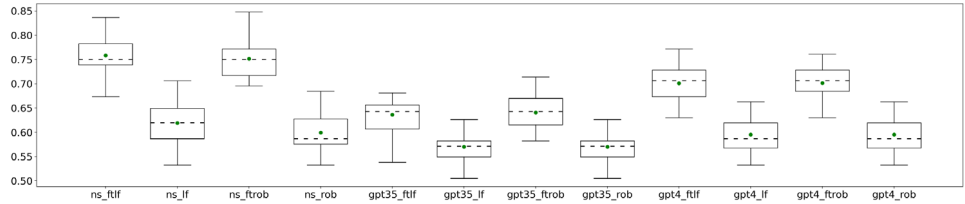
**Figure 5**: Accuracy of the classification with the CNN classifier

Figure 4 shows the boxplots for the F1 measures of 12 experiments that we have completed. The horizontal axis shows the codes for the experiment. The prefix "ns" indicates that the disputes were from the TWJY, and "gtp35" and "gpt4" indicate that the disputes were from GPT-3.5 and GPT-4, respectively. The suffix "ft" indicates that the BERT model was fine-tuned, "lf" indicates that the Lawformer was used for embedding, and "rob" indicates that the RoBERTa was used for embedding. The green dots and the orange segments in each boxplot mark the means and medians of the classification results, respectively. The vertical axis shows the F1 measures of the experiments. Putting the prefix and suffix together, "ns_ftlf" and "ns_ftrob" indicate that we fine-tuned the Lawformer and RoBERTa, respectively.

Using the Lawformer may provide better results than using the ordinary Chinese RoBERTa for us, but the differences were only obvious when we used the data in TWJY. The differences between the boxplots of "ns_lf" and "ns_rob" are more sensible than those between the boxplots of "gtp35_lf" and "gtp35_rob" and between "gtp4_lf" and "gpt4_rob".

We were surprised that the results of using the Lawformer did not necessarily outperform those of using the Chinese RoBERTa, no matter whether or not we fine-tuned them. We believe this is related to the fact that the Lawformer was built upon roberta-wwm-ext [17], and we used a larger and more recent version of RoBERTa (chinese-roberta-wwm-ext-large). After being fine-tuned with our data, the results of classifications improved significantly when we used the disputes in the TWJY in the experiments, as indicated by the boxplots "ns_ftlf", "ns_lf", "ns_ftrob", and "ns_rob".

The boxplots "ns_ftlf", "gtp35_ftlf", and "gtp4_ftlf" suggested two observations. The first is that the range of classification results shown in "ns_tflf" is much better than the results we reported in [10]. The second is very interesting: although the results of using the itemized disputes generated by GPT-3.5 and GPT-4 in the classifiers were inferior to those of using the disputes that were recorded in the original judgment documents. The differences were not very bad, and the gap shrank when we changed from GPT-3.5 to GPT-4. The same observations were repeated when we replaced the

Lawformer with the RoBERTa for embedding. Hence, it may become reasonable for us to use ChatGPT to summarize the litigants' disputes in our future work.

Figure 5 provides the boxplots for the accuracy of the 12 experiments. The qualitative observations remain the same.

## 7  Concluding Remarks and Further Discussions

We reported two lines of work in this paper. The first is about the classification procedure that consisted of a clustering component and a convolutional neural network component. This hybrid design outperformed our previous recommender for similar labor and employment cases [10].

Our second exploration may be more interesting for the JURISIN. We relied on the itemized disputes that the courts prepared for clustering and classification. Not all judgment documents we could download from TWJY contained such itemized disputes. The availability of such itemized disputes confines the scale of our experiments, so we tried to use the ChatGPT to summarize the claims and to itemize the disputes of the litigants. We designed our prompts to avoid possible hallucinations of the ChatGPT. We evaluated the GPT-generated disputes by using them in place of the court-prepared disputes. The results, as depicted in Figures 4 and 5, were quite encouraging. We observed that GPT-4 was better than GPT-3.5 in the final results. As we read the GPT-generated disputes in person, we also observed the superiority of GPT-4. As the technology of large language models has advanced so much recently, we look forward to the days when they become practically useful.

The world of LLMs, including ChatGPT, is changing at an extremely high speed. Since ChatGPT was launched in November 2022, GPT-4, GPT-4 Turbo (or GPT-4-1106-preview)[2], and GPT-4-0125-preview[3] followed within 14 months. The experimental results reported in this manuscript were based on GPT-3.5 and GPT-4. We have completed some preliminary evaluations of GPT-4 Turbo and GPT-3.5 Turbo for our tasks, but we have not evaluated GPT-4-0125-preview completely yet. Using GPT-4 Turbo and GPT-3.5 Turbo offered better results than using the previous GPT-4 and GPT-3.5. That was partially due to the increases in allowed tokens in the conversations. The improvements are not surprising and are certainly welcomed, as we have discussed in Section 6. The evaluation of the LLMs is not a trivial task, especially when the LLMs evolve so quickly and so aggressively [12]. We plan to investigate the factors that led to the improvements and analyze the influences of the factors on the effectiveness and efficiency of our clustering and classification components.

Reviewers of this paper recommended us to calculate the ROUGE scores [6] of the disputes that were generated by the LLMs unanimously, using the disputes that were listed in judgment documents as the references. ROUGE scores were used for evaluating the algorithmically generated summaries in the literature. However, there are different ways to the ROUGE scores, ROUGE-L in particular, when the generated summaries and the references have different numbers of sentences (itemized disputes in our work). In addition, we have to choose a segmentation tool for Chinese text. The task of evaluating the itemized disputes with ROUGE scores is not trivial, especially, given that an LLM may not generate the same disputes for repeated conversations, we should observe the distributions of the ROUGE scores.

---

[2] https://openai.com/blog/new-models-and-developer-products-announced-at-devday
[3] https://openai.com/blog/new-embedding-models-and-api-updates

|         | macro precision | | | macro recall | | | macro $F_1$ | | |
|---------|------|------|------|------|------|------|------|------|------|
|         | R-1  | R-2  | R-L  | R-1  | R-2  | R-L  | R-1  | R-2  | R-L  |
| GPT-3.5 | 0.530 | 0.275 | 0.373 | 0.321 | 0.149 | 0.190 | 0.372 | 0.175 | 0.221 |
| GPT-4   | 0.423 | 0.185 | 0.252 | 0.444 | 0.196 | 0.262 | 0.405 | 0.173 | 0.225 |

**Table 2.** Macro averages of the ROUGE scores for GPT-3.5-turbo and GPT-4 (0613)

We employed open-source tools for segmenting Chinese text[4] and calculating the ROUGE scores.[5] Since each judgement will have a set of ROUGE scores, so we computed the macro averages of the ROUGE scores for all of our judgment documents. At the time of writing, we have seen different versions of GPT-3.5 and GPT-4. The statistics in Table 2 are for the version of GPT of 13 June 2023.[6] Note that there are two possible definitions of macro $F_1$, and we adopted the ma$F_1$ in [14].

We have evaluated more recent GPTs, although we do not show their statistics here. Given that we observed that using GPT-4 to generate the itemized disputes for us generally led to better final results than using GPT-3.5 in Figure 4, we may tend to believe that the macro recall rates to evaluate the output of the GPTs is a better metric for our tasks.

**Acknowledgments**

We thank the reviewers for their valuable comments. This research was partly supported by project 110-2221-E-004-008-MY3 of the National Science and Technology Council of Taiwan. Po-Hsien Wu proposed and implemented the classifiers. Wei-Jie Lee handled the GPT-related tasks. Chao-Lin Liu led this and other cited projects, and wrote this paper. The authors discussed and cooperated on all research issues for the reported work.

## References

1. Cui Y, Che W, Liu T, Qin B, Wang S, and Hu G. Revisiting pre-trained models for Chinese natural language processing. *Proc. of the 2020 Conf. on Empirical Methods in Natural Language Processing*. 2020; p. 657–668.
2. Dan J, Xu L, and Wang Y. Integrating legal event and context information for Chinese similar case analysis. *Artificial Intelligence and Law*. October 2023; published online.
3. Fang J, Li X, and Liu Y. Low-resource similar case matching in the legal domain. *Lecture Notes in Computer Science 13530*. 2022; p. 570–582.
4. He T, Lian H, Qin Z, Zou Z, and Luo B. Word embedding based document similarity for the inferring of penalty. *Proc. of Int'l Conf. on Web Information Systems and Applications*. 2018; p. 240–251.
5. Kumar S, Reddy P K, Reddy V B, and Suri M. Finding similar legal judgements under

---

[4] These tools include `hfl/chinese-roberta-wwm-ext-large`, `AutoTokenizer`, and `SentenceTransformersTokenTextSplitter`.

[5] `rouge-chinese`: https://pypi.org/project/rouge-chinese/

[6] https://platform.openai.com/docs/models/gpt-4-and-gpt-4-turbo

common law system, *Proc. of the Eighth Int'l Workshop on Databases in Networked Information Systems*. 2013; p. 103–116.

6. Lin C-Y. ROUGE: A package for automatic evaluation of summaries, *Text Summarization Branches Out*, 2004; p. 74–81.
7. Liu C-L and Hsieh C-D. Exploring phrase-based classification of judicial documents for criminal charges in Chinese. *Lecture Notes in Artificial Intelligence* (LNAI) 4203. 2006; p. 681–690.
8. Liu C-L, Lin H-R, Liu W-Z, and Yang C. Functional classification of statements of Chinese judgment documents of civil cases (alimony for the elderly), *Proc. of the Thirty-Fifth Int'l Conf. on Legal Knowledge and Information Systems*, 2022; p. 206–212.
9. Liu C-L, Liu W-Z, Wu P-H, Huang S-c, and Huang H-C. Modeling the judgments of civil cases of support for the elderly at the district courts in Taiwan, *Proc. of the Seventeenth Int'l Workshop on Juris-Informatics*. 2023; p. 163–176. (An extended version will appear in an LNAI book)
10. Liu C-L and Liu Y-F. Some practical analyses of the judgment documents of labor litigations for social conflicts and similar cases, CEUR Workshop Proceedings 3423: *Proc. of the Third Int'l Workshop on Artificial Intelligence and Intelligent Assistance for Legal Professionals in the Digital Workplace*, the Nineteenth Int'l Conf. on Artificial Intelligence and Law. 2023; p. 100–109.
11. Mandal A, Ghosh K, Ghosh S, Mandal S. Unsupervised approaches for measuring textual similarity between legal court case reports. *Artificial Intelligence and Law*. 2021; 29:417–451.
12. Ma M and Mandal J. Overcoming Turing: Rethinking evaluation in the era of large language models, Stanford Law School Blogs, Nov. 16, 2023. https://law.stanford.edu/2023/11/16/overcoming-turing-rethinking-evaluation-in-the-era-of-large-language-models/
13. Reimers N and Gurevych I. Sentence-BERT: Sentence embeddings using siamese BERT-networks, *Proc. of the 2019 Conf. on Empirical Methods in Natural Language Processing and the Ninth Int'l Joint Conf. on Natural Language Processing*. 2019; p. 3982–3992.
14. Takahashi K, Yamamoto K, Kuchiba A, and Koyama T. Confidence interval for micro-averaged F1 and macro-averaged F1 scores, *Applied Intelligence*, 2022; 52:4961–4972.
15. Weiser B and Schweber N. The ChatGPT lawyer explains himself. *New York Times*. June 8, 2023. https://www.nytimes.com/2023/06/08/nyregion/lawyer-chatgpt-sanctions.html
16. Weiss D C. Latest version of ChatGPT aces bar exam with score nearing 90th percentile. *ABA Journal*. March 16, 2023. https://www.abajournal.com/web/article/latest-version-of-chatgpt-aces-the-bar-exam-with-score-in-90th-percentile
17. Xiao C, Hu X, Liu Z, Tu C, and Sun M. Lawformer: A pre-trained language model for Chinese legal long documents. *AI Open*. 2021; 2:79–84.
18. Xiao C, Zhong H, Guo Z, Tu C, Liu Z, Sun M, Feng Y, Han X, Hu Z, Wang H, and Xu J. CAIL2018: A large-scale legal dataset for judgment prediction. 2018; arXiv:1807.02478.
19. Zhang G, Lillis D, and Nulty P. Can domain pre-training help interdisciplinary researchers from data annotation poverty? A case study of legal argument mining with BERT-based transformers. *Proc. of the Workshop on Natural Language Processing for Digital Humanities*. 2021; p. 121–130.

## Appendix A

This appendix offers Chinese examples for what we discussed in Section 3.3. We show two dispute lists in Chinese. This appendix was copied from [10].

**Example 1**

The source is CHDV,92,勞訴,32,20040102,1.json.
1. 系爭夜點費應否列入平均工資核發退休金？
2. 原告己○○、丙○○、乙○○、戊○○、辛○○、庚○○於具領退休金時，所簽立之收據之效力為何？

**Example 2**

The source is CHDV,98,勞訴,37,20100409,1.json.
1. (一)長森醫院於97年7月31日是否有歇業之事實？
2. (二)兩造間是否因歇業而終止勞動契約？

After the blurring step (cf. Section 3.3), the anonymized person names in the second dispute in Example 1 were changed to "someone" in Chinese, which is listed below.

原告某人、某人、某人、某人、某人、某人於具領退休金時，所簽立之收據之效力為何？

The place name and the time expression in the first dispute in Example 2 were changed to "somewhere" and "sometime" in Chinese as well.

（一）某地於某時是否有歇業之事實？

# Appendix B

This appendix provides the Python code mentioned in Section 4. We translated the Chinese directives to English in the following illustration. The values of the two variables, i.e., `p_point` and `d_point`, would be replaced by the actual statements of the plaintiffs and the defendants for individual cases, respectively.

**#step1(summarizing the claim of plaintiff)**

```
prompt = f'''List the key points of the following article using {'p_point': []} as the template\n :{plaintiff_claim}'''
```

**#step2(summarizing the claim of defendant)**

```
prompt = f'''List the key points of the following article using {'d_point': []} as the template\n :{defendant_claim}'''
```

**#step3(dispute extracting)**

```
prompt = f'''The following are the main points argued by the claimant and the counterparty. Based on this, list the dispute points between the two parties, and use {'dispute': []} as the template\n The following is the plaintiff's claim:\n {p_point} \n The following is the defendant's claim:\n {d_point} '''
```